# Differential Evolution with Generalized Mutation Operator for Parameters Optimization in Gene Selection for Cancer Classification

Hossein Sharifi Noghabi[a,b], Habib Rajabi Mashhadi[a,b,c,1] and Kambiz Shojaei[a,b]

[a]Center of Excellence on Soft Computing and Intelligent Information Processing (SCIIP) located at Faculty of Engineering at Ferdowsi University of Mashhad, Iran.

[b]Department of Computer Engineering, Ferdowsi University of Mashhad, Iran.

[c]Department of Electrical Engineering, Ferdowsi University of Mashhad, Iran.

Abstract

Differential Evolution (DE) proved to be one of the most successful evolutionary algorithms for global optimization purposes in continuous problems. The core operator in DE is mutation which can provide the algorithm with both exploration and exploitation. In this article, a new notation for DE is proposed which has a formula that can be utilized for generating and extracting novel mutations and by applying this new notation, four novel mutations are proposed. More importantly, by combining these novel trial vector generation strategies and four other well-known ones, we proposed Generalized Mutation Differential Evolution (GMDE) that takes advantage of two mutation pools that have both explorative and exploitative strategies inside them. Results and experimental analysis are performed on CEC2005 benchmarks and the results stated that GMDE is surprisingly competitive and significantly improved the performance of this algorithm. Finally, GMDE is also applied to parameters optimization, modification and improvement of a feature selection method for cancer classification purposes over gene expression microarray profiles.

[1] h_mashhadi@um.ac.ir; hsharifi@stu.um.ac.ir

## I. INTRODUCTION

Evolutionary algorithms such as Genetic Algorithm (GA) or Evolutionary Strategy (ES) are stochastic search methods that are generally inspired from natural phenomena like natural selection or survival of the fittest [1, 2]. In recent years, evolutionary algorithms were applied to solve numerous real world problems and among these algorithms Differential Evolution (DE) has received so much attention in both developing itself and application areas [3]. DE was originally proposed in 1995 by Storn and Price for solving Chebyshev polynomial fitting problem [4]. DE has numerous advantages in comparison with other methods such as simple structure and implementation, low computational complexity, high performance, being robust and the fact that it is intersection of classic methods such as Controlled Random Search and evolutionary methods such as GA and Simulated Annealing (SA) [5]. The structure of its differential mutation operator and its crossover has made DE more suitable for continuous problems [3, 5]. More information is available in two good reviews [3, 6].

Generally, an evolutionary algorithm such as DE starts its optimization process with an initial population, this population is evaluated by cost/fitness function and better individuals will move to the next generation [7]. Through this process, individuals are evolved with mutation and crossover operators and this evolution will be continued until the algorithm reaches the acceptable answer or a stopping condition [3]. Mutation operator is utilized in order to generate a *donor* vector and after that the generated vector is used in crossover for generating a new trial vector or a candidate individual [8]. In fact, the goal of mutation operator is to provide the algorithm with more diversity, exploration and reliability, therefore, mutation is the core operator in DE [5]. In this operator, three vectors are chosen randomly, however, because of this randomness, one of the main drawbacks of DE is its weak exploitation capability [9]. The most important issue regarding evolutionary algorithms is how to establish a tradeoff between exploration and exploitation. Exploration or reliability helps the algorithm to explore all of the problem space to find promising regions and exploitations enhances the process of finding the optimum in that region [10]. Using inappropriate vector generation strategy or parameters setting can lead to over-explorative or over-exploitative behaviors just like random search or greedy search [11]. Although, an over-explorative mutation can determine most of the problem space but without exploitation pressure there would be no effective convergence. Therefore, providing the balance between exploration and exploitation is center of studies regarding most of the evolutionary algorithms such as DE [12]. Moreover, DE can have numerous applications [3, 6] ranged from engineering area to biological science [13].

This paper investigates whether the performance of DE can be improved by generalizing diverse mutations that have both explorative and exploitative properties. In this paper, we proposed a new notation for DE that can generates novel mutations for this algorithm and fills the empty places of the stated tradeoff. Synergism of these generated mutation and previous mutations in two mutation pools significantly improved the performance of DE in almost all of the cases.

Contributions of this paper can be summarized as follows:

- Proposing a new equation (notation) for DE mutation operator in order to extract and generate novel operators.
- Proposing four new mutation operators for DE which are obtained via utilizing the stated notation.
- Proposing a new ensemble approach for mutation based on eight operators divided into two mutation pools.
- Finally, we proposed Modified Maximum-Minimum Correntropy Criterion (M3C2). In this novel feature selection method GMDE is applied in a real-world problem for parameters optimization in feature (gene) selection for cancer classification.

The rest of the paper is organized as follow: Section II is about categorizing mutations in DE literature, Section III describes the proposed method and Section IV presents results and experimental analysis and finally Section V concludes the paper.

## II. RELATED WORK

Mutation is the core operator in DE and therefore, so many efforts are performed to improve and develop this operator [3]. In fact, mutation is the key to distinguish between variants of DE by applying DE/X/Y/Z notation where DE denotes differential evolution, X denotes the target or base vector, Y denotes number of randomly selected difference vectors and Z indicates type of crossover operator [3, 6]. Since in this paper we applied binomial crossover (bin), Z is omitted from this notation in further usage of it. In this section, we categorized previous efforts on mutation into four generations and discussed each of them briefly.

### A. First generation: Storn and Price mutations

In honor of the researchers that proposed DE for the first time, we entitled this generation Storn and Price. In this generation, mutations exist that have been widely used in DE literature so far. Most of these mutations are proposed by Storn and Price and they are following the classic notation of DE/X/Y. Some of the well-known mutations of this family are listed as follow [3, 5]:

DE/rand/1

$$X_{new,g} = X_{r1,g} + F \cdot (X_{r2,g} - X_{r3,g}) \tag{1}$$

DE/best/1

$$X_{new,g} = X_{best} + F \cdot (X_{r1,g} - X_{r2,g}) \tag{2}$$

DE/rand/2

$$X_{new,g} = X_{r1,g} + F \cdot (X_{r2,g} - X_{r3,g}) + F \cdot (X_{r4,g} - X_{r5,g}) \tag{3}$$

DE/best/2

$$X_{new,g} = X_{best} + F \cdot (X_{r1,g} - X_{r2,g}) + F \cdot (X_{r3,g} - X_{r4,g}) \tag{4}$$

DE/current-to-best/1

$$X_{new,g} = X_{current,g} + F \cdot (X_{best} - X_{current,g}) + F \cdot (X_{r2,g} - X_{r3,g}) \tag{5}$$

DE/rand-to-best/1

$$X_{new,g} = X_{r1,g} + F \cdot (X_{best} - X_{r1,g}) + F \cdot (X_{r2,g} - X_{r3,g}) \tag{6}$$

DE/current-to-rand/1  $(F' = ki \cdot F)$

$$X_{new,g} = X_{current,g} + k \cdot (X_{r1,g} - X_{current,g}) + F' \cdot (X_{r2,g} - X_{r3,g}) \tag{7}$$

Where, $X_{new,g}$ is the generated vector by mutation operator, $X_{best}$ indicates the best individual, $X_{current,g}$ indicates the current member of the population at generation $g$, $r_1 \neq r_2 \neq r_3 \neq r_4 \neq r_5 \neq current \neq new$ are randomly selected indices and $F$ is the scaling factor. Please note that in equations (1)-4) and (6) the *current* index is not present, however the $r_1 \ldots r_5$ need to be different from *new*.

Some of the mutations in this generation are explorative such as DE/rand/1 or DE/rand/2 and some others are more exploitative such as DE/best/1 or DE/current-to-best/1.

The main drawback of DE is its exploitation which means there is no adequate tradeoff in exploration and exploitation in DE because it explores the problem space more and the exploitation phase in it is not well enough [14]. The importance of this generation is the fact that, these mutations are foundation of DE and even today they are acting as criteria for evaluation of new variants of DE. The next generation is dealing with efforts to provide the algorithm with a balance between exploration and exploitation.

## B. Second generation: A quest for better tradeoff

In this family, the simple structure of the mutation operator according to its classic notation is deformed. This deformation occurred in hope of establishing a better tradeoff between exploration and exploitation. Mutation such as DE/rand/2/dir, trigonometric mutation, DE/rand/1/Either-or-algorithm and DEGL are some of the examples of this generation [12, 15, 16].

The other aspect of this generation is applying concepts such as niching and crowding in DE especially when it comes to multimodal problems [17]. In addition to fitness sharing approaches, in this generation, dividing population to sub-populations and performing diverse mutations on them can be considered another part of this generation [18-21].

The next generation is about an important step in evolution of every evolutionary algorithm.

## C. Third generation: Adaptive mutations

One of the most important developments on every evolutionary computation algorithm is parameters control and adaptation [22]. Controlling parameters proved to be significantly influential on general performance of an evolutionary computation method. Similar to other algorithms, DE is receiving a lot of attention regarding its parameters and how to control them [3]. Since scaling factor has impact on performance of mutation operator in DE directly, this generation is dealing with efforts and tries to control this parameter. This family of DE starts from deterministic parameter control and fuzzy logic [23] to adaptive and self-adaptive parameter control methods such as jDE [24], JADE [25], CoDE [26], SaDE [27], etc. which are well-known members of this generation.

It is important to note that algorithms such as SaDE and CoDE also take advantage of mutation pool. SaDE selects one mutation from its pool according to success and failure rates of that mutation in previous usages but CoDE performs all of the mutations in its pool and keeps the best obtained result from them.

The next generation is dealing with making mutation operator in DE intelligent in its selection of parent vectors.

## D. Fourth generation: Intelligent Selection

Usually, Parent vectors in DE are selected randomly from population, however, since good species in the population have good and valuable information, they are more likely to guide the other individuals in the population [14]. Based on this fact in nature, this generation tries to select parent vector intelligently by a meaningful criterion. Generally, this criterion is from fitness space such as Ranking-based mutation [14] or it is adopted from design space such Proximity-based mutation [28].

Another algorithms such as DERL [29], JADE [25], DE based on FER [30] can be considered other members of this generation.

The importance of this generation is the fact that this family provides DE with more exploitation which is helpful regarding the main drawback of it.

*E . An example of DE application in gene selection*

jDE has been recently applied for parameter optimization in Maximum-Minimum Correntropy Criterion (MMCC) for robust and stable gene selection for cancer classification over gene expression microarray profiles [13], In this article, the optimal value of three parameters related to this feature selection method including, sigma of the Correntropy, number of features and the regularization term for scoring each feature are determined by jDE (Figure 1 illustrates MMCC procedure).

In the next sections, first, GMDE is proposed and presented in details and after evaluation of its effectivity and optimality over numerical benchmarks, we applied it to improve MMCC and propose a modified version of it with much better performance.

## III. PROPOSED METHOD

In this section, we proposed a novel notation for DE which is capable of generating and extracting new mutation operators similar to the first generation mutations. The classic notation does not have such a capability and it is limitative. New mutations can be obtained by following equation:

$$V_i = X_{t_1} + \sum_{j=1}^{n} F_j * (X_{t_{2j}} - X_{t_{2j+1}}) \tag{8}$$

Where, $V_i$ is donor vector, $n$ is number of difference vectors, $F_j$ is the scaling factor of the *j-th* difference vector and $X_{ti}$ is a member of population. Determination of $X_{ti}$ is extremely related to the problem and purpose of the operator. There are numerous options for each of $X_{ti}$s in this equation including [3], best individual, current member, random member, top *p*% member [25], worst *p*% member, tournament winner (or any other selection method) [29], a member from a neighborhood [21] or an intelligently selected member according to fourth generation. Therefore, via playing with these building blocks and specifying each member a novel mutation can be obtained. For instance, if exploration is crucial for a problem, placing an explorative modules such as randomly selected members makes the generated mutation explorative or if exploitation is required, utilizing exploitative modules such as best or top *p*% makes the new mutation exploitative.

In this paper, all of the mutations for n = 1 and n = 2 are generalized and evaluated as a preprocess phase and then the best ones are selected for second round evaluation. In the second round, selected mutations take advantage of higher function evaluation and a self-adaptive parameter control mechanism. The best mutations in the second round are designated for Generalized Differential Evolution (GMDE) mutation pools. These top mutations are categorized into two mutation pools and at each generation only mutations of one of these pools are executed and the best generated individual will be the candidate for entering to the next generation.

In order to determine the mutation pool, a Set Selection Rate (SSR) parameter is defined. Designated mutations should be categorized into two pools in a way that both of the pools have explorative and exploitative mutations. This issue is important because it provides the algorithm with both exploration and exploitation for the entire process of optimization.

Therefore, the new notation for DE is presented as follow:

$$GMDE(T, F, d_1, d_2) \tag{9}$$

In this notation, *GMDE* denotes, Generalized Differential Evolution, F denotes types of scaling factor for instance in case of applying jDE parameter control for all of the difference vectors it would be *"jDE"* instead of *F* which size of *F* is $n*1$. For difference vectors, $d_1$ and $d_2$ are proposed and both of them are $n*1$. $d_1$ indicates the first element of all of the difference vectors and $d_2$ indicates the second elements of these vectors. *T* in the proposed notation, denotes the target or base vector of the mutation, in fact, $X_{t1}$ in Eq. (8) determines *T* in Eq. (9) and $X_{t2j}$ and $X_{t2j+1}$ are specifying difference vectors for Eq. (9). For instance, DE/rand/1 mutations in classic notation that lets use jDE as a parameter control is the following in the new notation:

*GMDE (rand,"jDE",rand,rand).*

Algorithm 1 is pseudocode of the GMDE.

Algorithm 1 GMDE:
1: Initialization of parameters such as SSR.
2: Evaluation the initial population.
3: **while** (termination criterion is not satisfied)
4:   select the mutation pool according to SSR.
5:   perform mutations of the selected pool.
6:   perform crossover for all of the generated donor vectors.
7:   select the best trial vector generated by crossover operator.
8:   **if** (trial vector < current member of the population) **then**
9:     enter the trial vector to the population.
10: **endif**
11: **end while**

IV. RESULTS AND EXPERIMENTAL ANALYSIS

Comprehensive and extensive experiments are performed in order to study the performance of the proposed notation and GMDE. In this section, first parameters settings and benchmark functions are introduced, then, the preprocess phase is discussed in detail. After preprocessing, second round evaluation is presented and finally the performance of GMDE is investigated in benchmark functions and a real-world application.

*A. Benchmark functions*

We utilized 25 benchmark functions from CEC2005 competition [31]. These functions are categorized into three groups: 1) F1-F5: these functions are unimodal, 2) F5-F12: these functions are basic multimodal functions, 3) F13 and F14 are expanded multimodal functions and finally 4) F15-F25 are hybrid composition functions.

*B. Parameters settings*

Parameters are extremely influential on the performance of the algorithm, in all cases, we applied jDE parameter control to have fair situation for all of the mutations and algorithms except when the algorithms has its own parameter control methods such as SaDE or CoDE. jDE [24] controls scaling factor and crossover rate as follow:

$$F_{i,G+1} = \begin{cases} F_l + rand_1 * F_u & rand_2 < \tau_1 \\ F_{i,G} & \text{otherwise} \end{cases} \quad (10)$$

$$Cr_{i,G+1} = \begin{cases} rand_3 & rand_4 < \tau_1 \\ Cr_{i,G} & \text{otherwise} \end{cases} \quad (11)$$

Where, $rand_{i,\ i \in 1...4} = \{rand \in [0,1]\}, \tau_1 = \tau_2 = 0.1,$
$F_l = 0.1, F_u = 0.9$

This approach makes $F \in [0.1, 0.9]$ and $Cr \in [0,1]$.

The other parameters are initialized as follow:

*NP* as number of population=50; Maximum run=50; *D* as number of dimensions=30 and maximum number of function evaluation=$D * 10\ 000$. These settings are applied for second round evaluation and GMDE phases and for preprocess phase another settings are used as follow:

*NP*=50; maximum number of function evaluation=10 000; *D*=10; Maximum run=100; *F*=0.5 and *Cr* as crossover rate=0.9;

*C. Preprocess phase*

In this section, diverse mutations generated by the proposed method are swept with stated settings in order to find potential mutations that can be useful for GMDE. In this phase, we set n=1 and n=2 but for simplicity only randomly selected member, the best member and current member of population are considered for blocks of the mutations. In addition to this assumption, in case of n=2, the second difference vector is also considered to be just random. The results for preprocess phase are presented in TABLE I through TABLE

VI. In these tables, $X_{ti}, i=1,2,...,5$ is determined according to Eq. (8) and effective mutations are specified in **boldface** and these mutations are considered for second round evaluation.

In case of n=1, mutations 1, 4, 6, 10, 12 and 16 have the best performance among other mutations. Mutation 1 was successful in 4 functions but was not very successful in others. Mutation 4 found the best answer in function 11 and achieved acceptable results in 10 other functions but was not entirely effective for functions such as 7, 8, 10 and 14. Mutation 6 obtained good results in functions 1 through 4 and also 6 and 16. Mutation 10 had the best result in functions 2, 5 and 6 and moreover, achieved good results in functions 1, 3, 4, 13 and 16. In comparison with other mutations, mutation 12 found the best outcome for functions 1, 4, 9 and 13 and also obtained good ones for functions 2, 3, 6 and 16. Finally, mutation 16 achieved the best result among other methods for 3 functions and also was successful for 6 other functions. These 6 mutations mostly were successful for unimodal functions and rarely effective for multimodal functions which was expectable due to lack of exploitation in DE. This drawback is significantly solved in mutations 4, 10 and 12 which take advantage of the best member in their structure. Interestingly, mutation 4 that has the best member in its difference vector appears to be strongly competitive in comparison with well-known mutation 10 (DE/best/1). Among these mutations, 1 and 10 were presented in DE literature before but 4, 6, 12 and 16 are suggested by this article.

In case of n=2, a second difference vector with randomly selected individuals is added to the mutation operators of n=1. This means that previous mutations are provided with more exploration and reliability. From 27 possible mutations in n=2, 11 mutations are selected for second round evaluation. Among them, 1, 4, 10, 21 and 24 were previously presented in other papers and mutations 11, 12, 13, 15, 16 and 17 are suggested in this paper. According to tables III and VI, using the current member of population is not achieving to competitive results. After determination of elite mutations, in the second round evaluation these mutations are provided with more function evaluation and parameter adaptation based on jDE as stated in the previous section. The results of second round evaluation are presented in TABLE VII and TABLE VIII for n=1 and n=2 respectively. In these tables results are evaluated by Wilcoxon Signed-rank test for $\alpha=0.05$ and according to this test $w$ indicates that the proposed mutation by Eq. (8) is significantly better than the compared mutation, $l$ indicates that the proposed mutation is worse and $t$ indicates that it is equal to the compared mutation.

For the second round evaluation of n=1, according to TABLE VII, the proposed mutations won in 73 cases, lost in 63 cases and performed equally in 13 cases. Mutation 4 obtained the best result comparing to mutations 1 and 10. This mutation won in 25, lost in 15 and tied in 5 comparing to previously suggested mutations (DE/rand/1 and DE/best/1). Mutation 12 in the same comparison won in 21, lost in 25 and tied in 4 cases and finally mutation 6 won in 22 lost in 24 and tied in 4 functions. Therefore, for mutations with three vectors

(one vector in the base and two in difference part) placing the best member of population in the first part of difference vector is leading to better results. In addition to this, applying current member in these family of mutations appears to be ineffective. Mutations 10 and 12 were not successful comparing to the other three mutation, however, they achieved to the best results for functions 3, 14 and 25. Mutation 1 was superior for hybrid functions (F15-F25) but it lost its superiority to mutation 4 in multimodal functions (F6-F14). Generally, the three proposed mutation (4, 6 and 12) were significantly better comparing to mutation 10 (DE/best/1) in both hybrid and multimodal functions, however, comparing to mutation 1 (DE/rand/1) they were better mostly in unimodal (F1-F5) and multimodal functions.

For second round evaluation of n=2, Mutations with 5 vectors (one vector in the base, and four ones in two difference vectors) are studied in this part of evaluation. According to TABLE VIII, mutations 1 and 4 were significantly better than proposed mutations for n=2. According to this table, among the proposed mutations, 12, 16 and 15 had the most wins, however, this performance was not well enough. From achieving the best result point of view, mutations 15 and 17 were the worst without any achievement and mutations 11 and 13 also only achieved the best results for two functions (functions 11 and for mutation 13 and function 14 for mutation 11). In order to save space, the details of the results for n=2 mutations are not stated. Both scenarios contain cases which are obviously ineffective such as mutation 5 in TABLE I or mutation 14 in TABLE III, however, we consider them in comparison in order to have a complete study.

*D. Generalized Mutation Differential Evolution (GMDE)*

So far 54 mutations were examined and evaluated in preprocess phase for n=1 and n=2 according to Eq. (8) and 16 of them were selected for the second round evaluation with stated parameters settings. According to the new notation in (9), these mutations are as follow (in "()" the old notation of those that presented before is stated):

1)* *GMDE#1(rand,jDE,rand,rand).* (rand/1)

2) *GMDE#2(best,jDE,rand,rand).* (best/1)

3) *GMDE#3(rand,jDE,best,current).*

4)* *GMDE#4(rand,jDE,best,rand).*

5) *GMDE#5(best,jDE,rand,current).*

6)* *GMDE#6(rand,jDE,$\frac{rand}{rand}, \frac{rand}{rand}$).* (rand/2)

7)* *GMDE#7(rand,jDE,$\frac{best}{rand}, \frac{rand}{rand}$).* (rand-to-best/1)

8) GMDE#8(best,jDE,$\frac{rand}{rand}$,$\frac{rand}{rand}$). (best/2)

9) GMDE#9(best,jDE,$\frac{rand}{best}$,$\frac{rand}{rand}$).

10)* GMDE#10(best,jDE,$\frac{rand}{current}$,$\frac{rand}{rand}$).

11)* GMDE#11(best,jDE,$\frac{best}{rand}$,$\frac{rand}{rand}$).

12) GMDE#12(best,jDE,$\frac{best}{current}$,$\frac{rand}{rand}$).

13)* GMDE#13(best,jDE,$\frac{current}{rand}$,$\frac{rand}{rand}$).

14) GMDE#14(best,jDE,$\frac{current}{best}$,$\frac{rand}{rand}$).

15)* GMDE#15(current,jDE,$\frac{rand}{current}$,$\frac{rand}{rand}$).(current-to-rand/1)

16) GMDE#16(current,jDE,$\frac{best}{current}$,$\frac{rand}{rand}$).(current-to-best/1)

Those 10 mutations that specified in boldface were able to find the best result for at least one function comparing to the other methods. Among them, 8 mutations that specified with "*" are selected for mutation pools. In the case of GMDE#9 and GMDE#3, although they were able to achieve the best outcome in 1 and 3 functions respectively, for GMDE#9 the standard deviation between all the mutations was near zero which means all the methods were performed almost the same and for GMDE#3, other mutations were able to achieve the same result as it did, consequently, in order to reduce these overlaps, GMDE#3 was not considered for mutation pools. The top 8 mutations are grouped into two mutation pools as follow:

$$\begin{cases} Pool\,1: GDE\,\#4, GDE\,\#6, GDE\,\#11\,and\,GDE\,\#15 \\ Pool\,2: GDE\,\#1, GDE\,\#7, GDE\,\#10\,and\,GDE\,\#13 \end{cases} \qquad (12)$$

Each pool contains both explorative and exploitative mutations and SSR determines the designated pool in each generation. In this article, being explorative or exploitative is simply defined by the structure of mutations. If a mutation has best or p% best components it is exploitative and if it has random modules it is explorative. SSR is initialized to 0.5, hence, both pools have 50% chance for selection as follow:

$$\begin{cases} if \quad rand \leq SSR \quad Select\,pool\,\#1 \\ if \quad rand > SSR \quad Select\,pool\,\#2 \end{cases} \qquad (13)$$

Where, $SSR = 0.5$ is set selection rate and $rand \in (0,1)$ is a randomly selected number.

The main advantage of such a method is synergism between old and new mutations and the fact that by applying the proposed notation, empty places are filled and competitive mutations such as GMDE#4 are obtained. At each generation, all of the mutations of the selected set are performed with different scaling factor which is adapted by jDE and each mutation strategy in the pool has different scaling factor. The resulted donor vectors are then applied in crossover but all of the crossover rates are the same for each donor vector and this rate is determined by jDE as well. The results of comparison between GMDE and similar methods are illustrated in TABLE IX. As stated in this table, with respect to other methods, in majority of functions, GMDE obtained significantly better value compared to the other methods. GMDE won in 18 functions in comparison with SaDE, only lost in 6 and tied in 1 function. Moreover, in comparison with CoDE, GMDE won in 15, lost in 7 and tied in 3 functions. In the case of unimodal and multimodal functions (F1-F14), GMDE completely overcomes the other methods and achieved the best results among them in 10 functions. In the case of hybrid functions, GMDE and CoDE, both found the best mean value in 5 functions and SaDE was successful in 4 functions. Moreover, comparing to other well-known DE variants, GMDE proved to be effective and competitive as presented in TABLE X. According to this table, GMDE overcomes, DERL, Ranking-based and Proximity-based mutations which means there is significant difference between obtained average performances of GMDE and corresponding method statistically in completely fair and equal conditions. In this table, MR- indicates mean of negative ranks, MR+ indicates mean of positive ranks, SR- denotes sum of negative ranks, SR+ denotes sum of positive ranks, P-value is the measure that determines the difference between algorithms is significant or not and finally in the last column, "+" indicates that GMDE is significantly better than the compared method, "-" indicates that GMDE is significantly worse than the compared method and "=" indicates that there is no meaningful difference between GMDE and the compared method. This method holds in the context of these experiments, under benchmark functions and settings used.

*E. Modified Maximum-Minimum Correntropy Criterion (M3C2)*

In this paper, we proposed M3C2 a modified version of MMCC that takes advantage of GMDE instead of jDE. The maximum accuracy of classification and optimal values of parameters are brought in Figure 2 and TABLE XI, respectively. As illustrated in this figure, the settings obtained by applying M3C2 achieved much better maximum accuracy in two data sets compared to the settings obtained by jDE and achieved the same results as jDE in two other ones and only was not able to outperform jDE settings in one of the studied data sets. Therefore, GMDE has significantly better performance than jDE in this investigated application. This superiority of GMDE settings is in the situations that for example, in TOX-171 data set the better performance is achieved by less number of features than jDE ones (see TABLE XI for more details).

The reason for GMDE performance is that it makes the best use of two mutation pools and as a result, it can explore and exploit the area of the problem more effectively.

## V. CONCLUSION AND FUTURE WORKS

In this article, we proposed a new notation for DE which provides us with mutation generation and extraction capabilities. By utilizing this notation, 57 mutations were examined in preprocess phase and 16 of them were chose for second round evaluation with more function evaluation and parameter adaptation by jDE. After the second round evaluation, those mutations that were able to find the best result among the others at least in one function were chose in order to generate mutation pools and these top 8 mutations were grouped into two mutation pools.

At each generation, only one of these pools is selected according to SSR parameter and all of the mutations in it are performed which means for each member of population 4 donor vector are generated. After crossover, the best trial vector among these 4 is chosen. The main advantage of GMDE is its synergism because of the fact that in each pool there are both explorative and exploitative mutations, hence, lack of exploitation in DE is solved. The results of extensive experiments on CEC2005 benchmarks stated that GMDE was significantly better than other methods with multiple mutations such as SaDE and CoDE and more importantly, GMDE also proved to be better than some other well-known DE variants such as DERL, Ranking-based and Proximity-based. For future works, investigation of the effect of population size and dimensionality on this method is of great importance and significance we also intend to work on adaptive GMDE for controlling SSR.

### ACKNOWLEDGMENT

The authors would like to thank Center of Excellence on Soft Computing and Intelligent Information Processing (SCIIP) for kind supports and Dr. Y. Wang for making the source code of jDE, SaDE and CoDE available online.

### CONFLICT OF INTERESTS

Authors have nothing to disclose regarding conflict of interests.